%% file: acl2021.tex
\documentclass[11pt,a4paper]{article}
\usepackage[hyperref]{acl2021}
\usepackage{times}
\usepackage{latexsym}

\usepackage{microtype}
\usepackage{colortbl}
\usepackage{times}
\usepackage{latexsym}
\usepackage{amsfonts}
\usepackage{graphicx}
\usepackage{caption}
\usepackage{subcaption}
\usepackage{url}
\usepackage{bm}
\usepackage{subfiles} 

\usepackage{booktabs}
\usepackage{amssymb}
\usepackage{amsmath}
\usepackage{enumitem}
\usepackage{mathtools}
\usepackage{multirow}
\usepackage{comment}

\definecolor{mygray}{gray}{0.9}

\newcommand{\change}[1]{{}#1}

\aclfinalcopy 

\title{Improving Counterfactual Generation for Fair Hate Speech Detection}

\author{Aida Mostafazadeh Davani,
        Ali Omrani,
        Brendan Kennedy,
        Mohammad Atari,\\
        \textbf{Xiang Ren,
        Morteza Dehghani} \\
    University of Southern California\\
    \texttt{\{mostafaz, aomrani, btkenned, atari, xiangren, mdehghan\}@usc.edu}  \\}

\begin{document}
\maketitle
\begin{abstract}
\vspace{-0.2cm}
Bias mitigation approaches reduce models' 
dependence on sensitive features of data, such as social group tokens (SGTs), resulting in equal predictions across the sensitive features.
In hate speech detection, however, equalizing model predictions may ignore important differences among targeted social groups, as hate speech can contain stereotypical language specific to each SGT.
Here, to take the specific language about each SGT into account, we rely on \textit{counterfactual fairness} and equalize predictions among counterfactuals, generated by changing the SGTs. 
Our method evaluates the similarity in sentence likelihoods (via pre-trained language models) among counterfactuals, to treat SGTs equally only within interchangeable contexts.
By applying logit pairing to equalize outcomes
on the restricted set of counterfactuals for each instance, we improve fairness metrics while preserving model performance on hate speech detection.
\end{abstract}


\section{Introduction}
\vspace{-0.1cm}
Hate speech classifiers have high false-positive error rates in documents mentioning specific social group tokens (SGTs; \textit{e.g.,} ``\textit{Asian}'', ``\textit{Jew}''), due in part to the high prevalence of SGTs in instances of hate speech \citep{wiegand2019detection, mehrabi2019survey}. When propagated into social media content moderation, this \textit{unintended bias} \citep{dixon2018measuring} leads to unfair outcomes, \textit{e.g.,} mislabeling mentions of protected social groups as hate speech.

\begin{figure}[t]
    \centering
    \includegraphics[width=.49\textwidth]{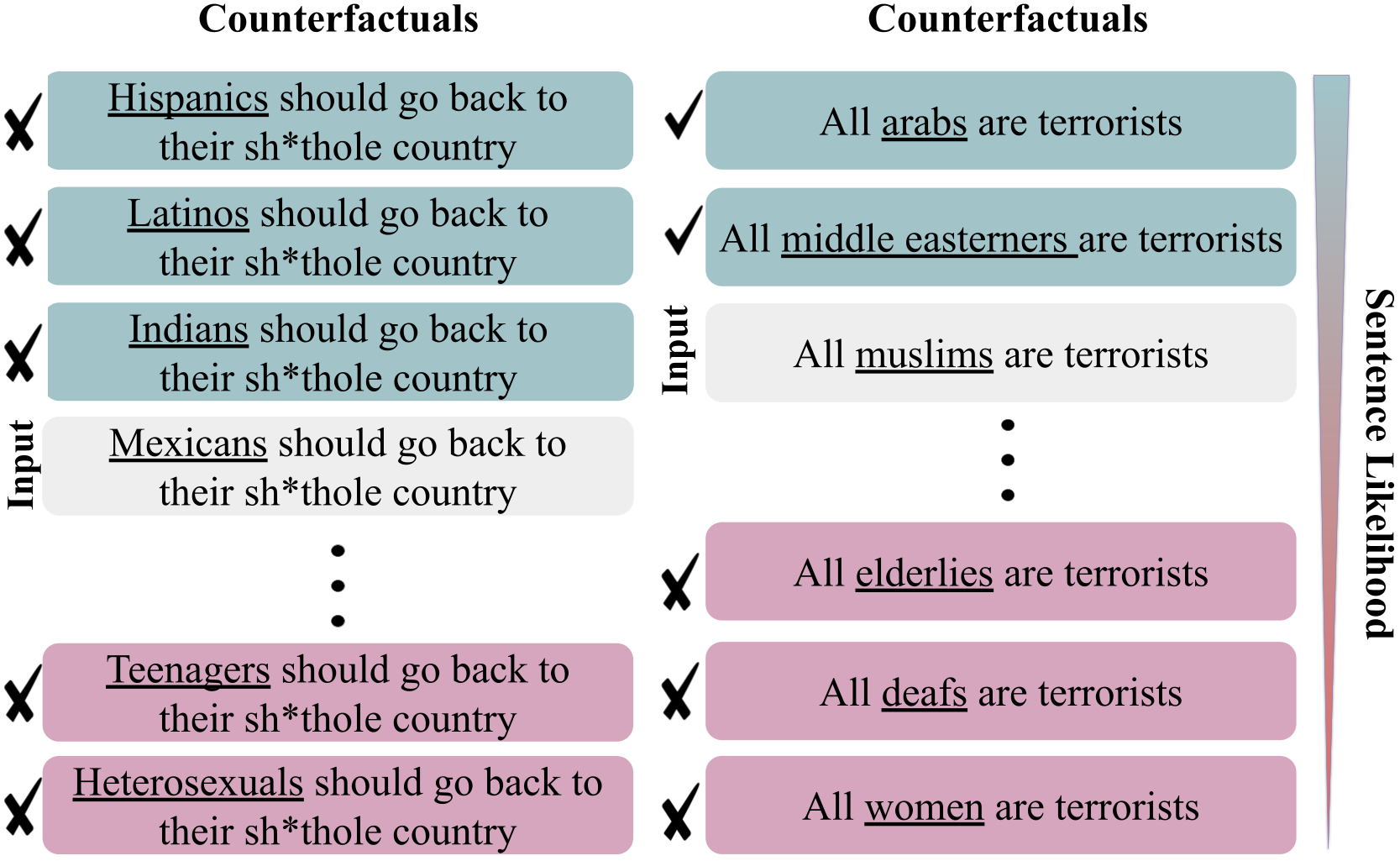}
    \caption{Two input sentences and their counterfactuals ranked by likelihood. Our method ensures similar outputs only for counterfactuals with higher likelihood.
    }
    \label{fig:counterfactual}
\end{figure}

For prediction tasks in which SGTs do not play any special role (e.g., in sentiment analysis), 
unintended bias can be reduced by optimizing group-level fairness metrics such as \textit{equality of odds}, which statistically equalizes model performance across all social groups \citep{hardt2016equality, dwork2012fairness}. However, in hate speech detection, this is not the case, with SGTs providing key information for the task (see Fig. \ref{fig:counterfactual}). 
Instead, bias mitigation in hate speech detection benefits from relying on individual-level fairness metrics such as \textit{counterfactual fairness}, 
which assess the output variation resulting from changing the SGT in individual sentences \citep{garg2019counterfactual, kusner2017counterfactual}.
Derived from causal reasoning, a counterfactual applies the slightest change to the actual world to assess the consequences in a similar world \citep{stalnaker1968theory, lewis1973counterfactuals}. 

Accordingly, existing approaches for reducing bias in hate speech detection using counterfactual fairness learn robust models whose outputs are not affected by changing the SGT in the input \citep{garg2019counterfactual}. 
However, a drawback of such approaches is the lack of semantic analysis of the input to identify whether changing the SGT leads to a small enough change that preserves the hate speech label \citep{kasirzadeh2021use}. For instance, in a hateful statement,  ``\textit{mexicans should go back to their sh*thole countries}'',
substituting  ``\textit{mexicans}'' with ``\textit{women}'' changes the hate speech label, while using ``\textit{Hispanics}'' should preserve the output. 
Here, we aim to create counterfactuals that \textit{maximally} preserve the sentence 
and  disregard counterfactuals that violate the requirement for being the ``\textit{closest possible world}'' (Fig.~\ref{fig:counterfactual}). 

To this end, we develop a counterfactual generation method 
which filters candidate counterfactuals based on their difference in likelihood from the actual sentence, estimated by a pre-trained language model with known stereotypical correlations \citep{sheng2019woman}. Intuitively, our method provides outputs that are robust with regard to the context and are not causally dependent on the presence of specific SGTs.  This use of sentence likelihood is inspired by \citet{nadeem2020stereoset} as it captures the similarity of an SGT and its surrounding words to prevent unlikely SGT substitutions. 
As a result, only counterfactuals with equal or higher likelihoods compared with the original (``closest possible worlds'') are used during training. To enforce robust outputs for similar counterfactuals, we apply logit pairing \citep{kannan2018adversarial} on outputs for sentence-counterfactual pairs, adding their average differences to the classification loss.  
Our method (1) effectively identifies semantically similar counterfactuals and (2) improves fairness metrics while preserving classification performance, compared with other strategies for generating counterfactuals.



\section{Related Work}
\vspace{-0.2cm}


Unintended bias in classification is defined as differing model performance on subsets of datasets that contain particular SGTs \citep{dixon2018measuring, mehrabi2019survey}.
To mitigate this bias, data augmentation approaches are proposed to create balanced labels for each SGT or to prevent biases from propagating to the learned model \citep{dixon2018measuring, zhao2018gender, park2018reducing}. 
Other approaches apply regularization of post-hoc token importance \citep{kennedy2020posthoc}, or adversarial learning for generating fair representations \citep{madras2018learning,zhang2018mitigating}
to minimize the importance of protected features.


By altering sensitive features of the input and assessing the changes in the model prediction, counterfactual fairness \cite{kusner2017counterfactual} seeks causal associations between sensitive features and other data attributes and outputs. Similarly, counterfactual token fairness applies counterfactual fairness to tokens in textual data \citep{garg2019counterfactual}.

Counterfactual fairness presupposes that the counterfactuals are close to the original world. 
However, previous work has yet to quantify this similarity in textual data. Key to our proposed framework 
is evaluating the semantic similarity between the original and the synthetically generated instances to only consider counterfactuals that convey similar sentiment. Consequently, our method prevents synthetic counterfactuals unlikely to exist in real-world samples which (1) decrease classification accuracy by adding noise into the training process and (2) misdirect fairness evaluation by introducing unexpected criteria.


\section{Method}
\label{sec:context}
\vspace{-0.2cm}



We propose a method for improving individual fairness in hate speech detection by considering the interchangeable role of SGTs in each specific context.
Given instance $x \in X$, and a set of SGTs $S$, we seek to equalize outputs of a classifier $f$ for $x$ and its counterfactuals $x_{cf}$ generated by substituting the SGT mentioned in $x$.

First, we provide the definition of \textit{counterfactual token fairness} (CTF), which can be evaluated for a model over a dataset of sentences and their counterfactuals (Sec.~\ref{subsec:definition}). Next, we specify how \textit{counterfactual logit pairing} (CLP) regularizes CTF in a classification task (Sec.~\ref{sec:clp}). Lastly, we introduce our counterfactual generation method for Assessing Counterfactual Likelihoods (ACL, Sec.~\ref{sec:gen_count}), which is driven by linguistic analysis of stereotype language in sentences. 


\subsection{Counterfactual Token Fairness (CTF)}
\label{subsec:definition}
Given instance $x \in X$, and a set of counterfactuals $\bm{x}_{cf}$, generated by perturbing mentioned SGTs, the CTF for a classifier $f = \sigma(g(x))$ is:
\begin{equation*}
    \mathrm{CTF}(X, f) = \sum_{x \in X}\sum_{x' \in x_{cf}}|g(x) - g(x')| 
\end{equation*}
where $g(x)$ returns the logits for $x$ \citep{garg2019counterfactual}. Lower CTF indicates similar (i.e., fairer) outputs for sentences and their counterfactuals.

\subsection{Counterfactual Logit Pairing (CLP)}
\label{sec:clp}


To reduce CTF while training a hate speech classifier, we apply counterfactual logit pairing (CLP) \citep{kannan2018adversarial} to all instances and their counterfactuals. 
CLP penalizes prediction divergence among inputs and their counterfactuals by adding the average absolute difference in logits of the inputs and their counterfactuals to the training loss: 
\begin{equation*}
      \sum_{x \in X} \ell_{c}(f(x), y) + \lambda\sum_{x \in X}\sum_{x' \in x_{cf}}|g(x) - g(x')|
\end{equation*}
where $\ell_{c}$ calculates the classification loss for an output $f(x)$ and its correct label $y$  and $\lambda$ tunes the influence of the counterfactual fairness loss, the impact of which is discussed in the Appendix.



\subsection{Counterfactual Generation}
\label{sec:gen_count}
Rather than simplifying the model training by restricting CLP loss to all counterfactuals created by perturbing the SGTs in non-hate sentences \citep{garg2019counterfactual}, we identify similar counterfactuals based on likelihood analysis of each sentence. Our aim is to generate counterfactuals that preserve the likelihood of the original sentence.

In stereotypical sentences that target specific social groups, expecting equal outputs when changing the SGT leads to ignoring how specific vulnerable groups are targeted in text \citep{haas2012hate}. Quantifying the change in a sentence as a result of perturbing SGTs has already been studied for detecting stereotypical language \citep{nadeem2020stereoset}; 
similarly to \citeauthor{nadeem2020stereoset}, 
we apply a generative language model \citep[GPT2;][]{radford2019language} to evaluate the change in sentence likelihood caused by substituting an SGT --- e.g., we expect the language model to predict decrease in likelihood for a sentence about terrorism when it is paired with ``\textit{Muslim}'' or ``\textit{Arab}'' versus other SGTs. 

Since GPT-2 uses the left context to predict the next word, for each word $x_i$ in the sentence, the likelihood of $x_i$, $P(x_i|x_0\ldots x_{i-1})$, is approximated by the softmax of $x_i$ with respect to the vocabulary. Therefore, the log-likelihood of a sentence $x_0, x_1, \ldots x_{n-1}$ is computed with:
$\lg P(x) = \sum_{i}^n \lg P(x_i|x_0,..,x_{i-1})$

\subfile{tables/SGT_tab.tex} 

We identify correct counterfactuals by comparing their log-likelihood to that of the original sentence and create the set of all correct counterfactuals $\bm{x}_{cf}$ by including counterfactuals with equal or higher likelihood compared with $\bm{x}$:
\begin{equation*}
x_{cf} = \{x' | x' \in \text{substitute}(x, S), P(x) \leq P(x') \}
\end{equation*}
in which substitute$(x, S)$ creates the set of all perturbed instance by substituting the SGT in $x$, with another SGT from the list of all SGTs $S$, which in this paper is a list of 77 SGTs (see Appendix), compiled from \citet{dixon2018measuring} and extended using 
WordNet synsets \citep{fellbaum2012wordnet}.
\subfile{tables/annotation.tex}


\subfile{tables/ghc_storm.tex} 



\section{Experiments}
Here, we apply our method for generating counterfactuals (Sec.~\ref{sec:gen_count}) to a large corpus to explore the method's ability to identify similar counterfactuals.
Then, we apply CLP (Sec.~\ref{sec:clp}) with different strategies for counterfactual generation and compare them to our approach, introduced in Sec.~\ref{sec:gen_count}.

\vspace{-0.1cm}
\subsection{Evaluation of Generated Counterfactuals}
\label{sec:exp_stereo}

\smallskip
\noindent
\textbf{Data.} We randomly sampled 15 million posts from a corpus of social media posts from Gab \citep{pushshift_gab}, and selected all English posts that mention one SGT (\textit{N} $\approx$ 2M).
The log-likelihood of each post and its candidate counterfactuals were computed. The primary outcome was the original instance's \textit{rank} in log-likelihood amongst its counterfactuals. Higher rank for a mentioned SGT indicates the stereotypical content of the sentence.

We conducted two qualitative analyses with human annotators to evaluate the generated counterfactuals. First, we selected sentences in which the highest ranks were assigned to the original SGTs and asked annotators to predict the mentioned SGT in a fill-in-the-blank test. If our method correctly ranks SGTs based on the context, we expect annotators to predict the original SGTs in such sentences. Then, we randomly selected a set of sentences and evaluated our method on finding the preferable counterfactual among a pair of candidates by comparing the choices to those of the annotators'.

Human annotators were from the authors of the paper, with backgrounds in computer science and social science. All annotators had previous experience with annotating hate speech content. However, they did not have any experience with the exact sentences in the evaluated dataset, given that the sentences were randomly selected from a dataset of 1.8M posts, collected by other researchers cited in the paper. 

We preferred expert annotators over novice coders in this specific case, because previous studies have indicated  expert coder higher performance in hate speech annotation \cite{waseem2016you}. Moreover, annotators' cognitive biases and perceived stereotypes can greatly impact their judgments in detecting hate speech \citep{sap2019risk}. Therefore, we preferred to have expert annotators with a shared understanding of the definition of stereotypes and hate speech, who are consequently less subjective in their judgments.

\smallskip
\noindent
\textbf{Results.} 
In 2.9\% of sentences the original SGT achieves the highest ranking. In 86.03\% of the posts where the original SGT is ranked second, the top-ranked SGT is from the same \textit{social category} (e.g., both SGTs referred to race or gender). 
We randomly selected 500 original posts with highest likelihood among their counterfactuals (Table~\ref{tab:stereo} shows such samples) to qualitatively assess their stereotypicality in a fill-in-the-blank style test with human subjects.  
Three annotators, on average, identified the correct SGT from 4 random choices for 74.88\% of posts. 
In a second evaluation, given sentences and two counterfactuals, annotators were asked to identify which SGT substitution preserves the hate speech and likelihood of the sentence. On average, annotators agreed with the model's choice in 63.07\% of the test items. Table \ref{tab:anno} demonstrates accuracy and agreement scores of annotators.


\vspace{-0.1cm}
\subsection{Fair Hate Speech Detection}
\label{sec:exp_hate}
\vspace{-0.1cm}
We apply our counterfactuals generation method to hate speech detection, and equalize model outputs for sentences and their similar counterfactuals.

\smallskip
\noindent
\textbf{Compared Methods.} We fine-tined BERT \citep{delvim2019bert} classifiers with CLP loss, using five approaches for generating counterfactuals: 1) \textbf{CLP+ACL} applies our approach for Assessing Counterfactual Likelihoods (Sec.~\ref{sec:gen_count}), 2) \textbf{CLP+NEG} considers all counterfactuals for negative instances \citep{garg2019counterfactual}, 3) \textbf{CLP+SG} substitutes SGTs from the same social categories (inspired by Sec.~\ref{sec:exp_stereo}), e.g., it replaces a racial group with other racial groups, 4) \textbf{CLP+Rand} substituting SGTs with random words, and 5) \textbf{CLP+GV} substitutes SGTs with ten most similar SGTs based on their GloVe word embeddings \citep{pennington2014glove}. As baseline models we consider a vanilla fine-tuned BERT (\textbf{BERT}), and a fine-tuned BERT model that masks the SGTs (\textbf{MASK})\footnote{Implementation details are provided in the Appendix}. 

\smallskip
\noindent
\textbf{Data.} We trained models on the Gab Hate Corpus \citep[\textbf{GHC};][]{kennedy2020gab} and Stormfront dataset \citep[\textbf{Storm};][]{de2018hate}, including approximately 27k and 11k social media posts respectively. Both datasets are annotated based on typologies that define hate speech as targeting individuals or groups based on their group associations.

\smallskip
\noindent
\textbf{Evaluation Metrics.} We compute CTF on two datasets of counterfactuals. (1) Similar Counterfactuals (SC; collected from \citet{dixon2018measuring}) includes synthetic, non-stereotypical instances based on templates (e.g., \textless You are a \verb|ADJ| \verb|SGT|\textgreater{}). In such instances, the sentence is not explicit to the SGT, and the model prediction should solely depend on the \verb|ADJ|s so smaller values of CTF are indicative of a fairer models. (2) Dissimilar Counterfactuals (DC; from \citet{nadeem2020stereoset}) includes stereotypical sentences and their counterfactuals generated by perturbing SGTs. Since instances are stereotypical, we expect all counterfactuals to be ignored by a fair model and lower CTF scores.

We also report group fairness metrics (equality of odds).
The standard deviation of true positive (TP) and true negative (TN) rates across SGTs are reported for a preserved test set (20\% of the dataset) and instances generated by perturbing the SGTs. The standard deviation of false positive ratio (FPR) for different SGTs are also reported for a dataset of non-hateful New York Times sentences. Lower standard deviations indicate higher group fairness.

\smallskip
\noindent
\textbf{Results.} Table \ref{tab:results_ghc_storm} shows the results of these experiments on  \textbf{GHC} and \textbf{Storm}. Evidently, our model (highlighted in Table \ref{tab:results_ghc_storm}) for generating counterfactuals enhances CTF while improving or preserving classification performance and group fairness (TP, TN, and FPR) on both datasets. The increase in classification performance demonstrates our method's capability in filtering noisy synthetic samples. 
These results call for further explorations of when fair models should treat SGTs equally. Rather than expecting equal results over all instances, fair predictions should be based on contextual information embedded in the sentences.



\vspace{-0.1cm}
\section{Conclusion}
\vspace{-0.2cm}

Our method treats social groups equally only within interchangeable contexts by applying logit pairing on a restricted set of counterfactuals.  We demonstrated that biased pre-trained language models could enhance counterfactual fairness by identifying stereotypical sentences. Our method improved counterfactual token fairness and classification accuracy by filtering unlikely counterfactuals. 
Future work may explore semantic-based techniques for creating counterfactuals in domains other than hate speech detection, e.g., crime prediction, to better contextualize definitions of social group equality. 

\subfile{ethics.tex}

\section*{Acknowledgments}
This research was sponsored in part by NSF CAREER BCS-1846531 to Morteza Dehghani.

\bibliographystyle{acl_natbib}
\bibliography{anthology,acl2021}

\clearpage

\appendix

\subfile{appendix.tex}

\end{document}

%% file: tables/SGT_tab.tex
\begin{table}[t]
\vspace{-0.1cm}
    \centering
    \scalebox{0.75}{
    \begin{tabular}{p{.6\textwidth}}
    \toprule
       \textbf{Stereotypical Sentences (from Gab)}\\
        \midrule
        \textbf{Communists} and dictators are desperate to get rid of god. His blessing overcomes the fearful evils of this fallen world.\\ \midrule 
        Dumb ass n**** don't realize you actually have to work your ass off on a farm. It doesn't just magically happen now that they've stolen the land from \textbf{Whites}.\\\midrule 
        Israel and the Islamist conspiracy to deny \textbf{Jews} their land. \\\midrule
        \textbf{Women}.  lie. about. rape.
        \\\bottomrule
    \end{tabular}
    }
    \vspace{-0.2cm}
    \caption{\small \textbf{Sample stereotypical sentences} from Gab, for which changing the SGT (bolded) decreases the likelihood.}
    \label{tab:stereo}
    \vspace{-0.1cm}
\end{table}

%% file: tables/annotation.tex
\begin{table}[t]
    \centering
    \scalebox{0.75}{
    \begin{tabular}{ccccc}
    \toprule
        Rank & \# Items & \# Choices & Accuracy(mean) & Agreement  \\
        \midrule
        1 & 500 & 4 & 74.88\% & 58.43\\
        \textgreater 1 & 250 & 2 & 63.07\% & 70.81\\
    \bottomrule
    \end{tabular}}
    \vspace{-0.2cm}
    \caption{\small Annotators' averaged accuracy and   agreement \citep{fleiss1971measuring} on sentences with different likelihood rankings.}
    \label{tab:anno}
    \vspace{-0.2cm}
\end{table}

%% file: tables/ghc_storm.tex
\begin{table*}[t]
    \centering
    \scalebox{0.78}{
    \begin{tabular}{l|c|ccc|cc|c|ccc|cc}
    \toprule
    \multicolumn{1}{c}{}& \multicolumn{6}{|c|}{\textit{GHC}} & \multicolumn{6}{c}{\textit{Storm}} \\\cline{2-13}
     \multicolumn{1}{c|}{} & \textbf{Hate} & \multicolumn{3}{c|}{\textbf{EOO}} & \multicolumn{2}{c|}{\textbf{CTF}} & \textbf{Hate} & \multicolumn{3}{c|}{\textbf{EOO}} & \multicolumn{2}{c}{\textbf{CTF}}   \\
        
        \multicolumn{1}{c|}{}& F1($\uparrow$) & TP($\downarrow$) & TN($\downarrow$) & FPR($\downarrow$) & DC($\downarrow$) & SC($\downarrow$) & F1($\uparrow$) & TP($\downarrow$) &  TN($\downarrow$) & FPR($\downarrow$) & DC($\downarrow$) & SC($\downarrow$) \\\midrule
        \textbf{BERT} & 
        73.30$\pm{.2}$ & 38.3 & 23.0 & 6.6 & 2.22 & 1.99 &  78.52$\pm{.2}$ & \textbf{40.8}& 25.3 & 11.5& 0.96& 1.16 \\
        \textbf{MASK}  & 
        71.24$\pm{.2}$ & 39.0& 20.3& 2.5 & 1.78 & 1.99 & 70.91$\pm{.2}$ & 43.4& 25.9& 8.3 & 0.96 & 1.16 \\
        \textbf{CLP+SG} & 
        62.10$\pm{.2}$ & 38.4& 23.2& \textbf{0.2}& 0.97& 2.24 &  80.31$\pm{.2}$ & 41.8& 25.4& 9.8& 0.71& 1.06\\
        \textbf{CLP+Rand} & 
        66.45$\pm{.2}$ & 41.3 & 20.4 & 2.7 & 0.98 & 1.24 & 80.62$\pm{.2}$ & 43.7 & 25.8 & \textbf{1.6} & 0.83 & 0.99 \\
        \textbf{CLP+GV} & 
        68.50$\pm{.2}$ & 38.3 & 23.0 & 3.1 & 1.01 & 1.25 & 79.28$\pm{.2}$ & 40.7 & 30.6 & 3.4 & 0.76 & 0.93  \\
        \textbf{CLP+NEG} & 
        70.02$\pm{.2}$& 39.6 & \textbf{20.1} & 7.7 & 0.76 & 1.98 &  77.62$\pm{.2}$& 42.5 & 26.2 & 5.0 & 0.56 & 0.98 \\
        
        \rowcolor{mygray} \textbf{CLP+ACL} & 
        \textbf{73.31$\pm{.2}$} & \textbf{37.5}& 20.5& 2.4 & \textbf{0.75}& \textbf{0.87}& \textbf{81.99$\pm{.2}$} & 42.8 & \textbf{23.1}& 2.0 & \textbf{0.42} & \textbf{0.53}\\\bottomrule
       
    \end{tabular}
    }
    \vspace{-0.1cm}
    \caption{\small \textbf{Results on GHC and Storm.} Baseline BERT model, and fine-tuned BERT masking SGTs, and five
    counterfactual logit pairing models (CLP) with counterfactual generation based on similar social groups (CLP+SG), random word substitution (CLP+Rand), GloVe similarity (CLP+GL), baseline approach \citep[CLP+NEG;][]{garg2019counterfactual}, and \setlength{\fboxsep}{0pt}\colorbox{mygray}{our approach for Assessing Counterfactual Likelihoods (CLP+ACL)}, trained in 5-fold cross validation and tested on 20\% of the datasets. Group-level fairness (true positive, true negative and false positive ratio) and counterfactual fairness are evaluated.} 
    \label{tab:results_ghc_storm}
\end{table*}

%% file: ethics.tex
\subsection*{Broader Impact Statement}
Our paper investigates bias mitigation in hate speech detection. This task is of great sensitivity because of the impact of online hate speech on minority social groups. While most discussions in the field of Ethics of AI focus on equalizing biases against different social groups from pre-trained language models, we make use of this bias to identify stereotypical or conspiratorial hate speech in social media and to ensure that hate speech detection models learn these linguistic association of stereotypes for protecting social groups from rhetoric that is explicitly targeting them.

%% file: appendix.tex
\section{Appendix}
All data is uploaded to dropbox\footnote{\url{https://www.dropbox.com/s/awjvtt5op43ewr6/Data.zip?dl=0}}
\subsection{Social Group Tokens}
\label{app:sgt}
Our social group terms include:
heterosexual, catholic, queer, latinx, younger, christian, latin american, jewish, jew, democrat, republican, indian, trans, canadian, white, bisexual, female, men, man, women, woman, gay, paralytic, blind, aged, spanish, taiwanese, taoist, protestant, paralyzed, liberal, deaf, buddhist, chinese, african, older, elder, deafen, latino, straight, latina, english, asian, male, amerind, old, american, conservative, japanese, muslim, homosexual, nonbinary, lesbian, protestant, ashen, sikh, lgbt, teenage, middle eastern, hispanic, bourgeois, lgbtq, european, millenial, transgender, african, young, elderly, paralyze, middle aged, black, mexican, arab, immigrant, migrant, and communist
\subsection{Study 1: Qualitative Analysis}
\textbf{Implementation Details}
To compute perplexity scores of the counterfactuals, we used 41 Google cloud virtual machine instances with the following configuration. All the instances used the Google n1-standard-4 (4 vCPUs, 15 GB memory). We had 1 x NVIDIA Tesla P100 Virtual Workstation, 15 x NVIDIA Tesla P4 Virtual Workstation, and 25 x NVIDIA Tesla K80. In addition we used one local machine with 1 x NVIDIA GeForce RTX 2080 SUPER, AMD Ryzen Threadripper 1920X CPU and 128 GB memory.
    
For each data point, the runtime of generating 64 counterfactuals along with their perplexity scores was about 1.5 seconds on instances with NVIDIA Tesla P4 Virtual Workstations and NVIDIA GeForce RTX 2080 SUPER and about 2.6 seconds on instances with NVIDIA Tesla K80 GPUs.
    
\textbf{Hyper parameters}
We used the pre-trained GPT-2 model from the transformers library by hugging face \footnote{\url{https://huggingface.co/transformers/pretrained_models.html}} with 12-layer, 768-hidden, 12-heads, 117M parameters.
    
\textbf{Dataset}
We downloaded the public dump of Gab posts \footnote{\url{https://files.pushshift.io/gab/}} which contains more than 34 million posts from August 2016 to October 2018. After dropping posts with small number of English tokens (non-English posts) and malformed records, We got near 15 million posts referred to as \textbf{SGT-Gab}. Data can be found in the accompanied zip file.



\subsection{Study 2}

\textbf{Implementation Details}
Each of the seven models were trained on 80\% of the given dataset (either \textbf{GHC} or \textbf{Storm}), (\textit{dataset}\_train.csv file) and tested on the remaining 20\% (\textit{dataset}\_test.csv file). The models were run on a single NVIDIA GeForce GTX 1080 GPU, where each epoch takes 3 seconds. 
Models were built in Python 3.6 and Tensorflow-GPU 
\citep{abadi2016tensorflow}.


Data cleaning was performed by applying the \textbf{BertTokenizer} tokenizer \citep{wolf-etal-2020-transformers}, 
and models were trained by fine-tuning \textbf{Bert-For-Sequence-Classification} initialized with pre-trained ``bert-base-uncased''\footnote{\url{https://huggingface.co/bert-base-uncased}} with 12-layers, 768-hidden, 12-heads, and 117M parameters \citep{wolf-etal-2020-transformers}. The $\lambda$ coefficient was set to 0.2 for all models to specify the same counterfactual loss in all models.

\textbf{Hate Speech Datasets}
Here we provide detail on the two training datasets from our experiments. 
The Gab Hate Corpus \citep[\textbf{GHC};][]{kennedy2020gab} is an annotated corpus of English social media posts from the far-right network ``Gab.'' Labels were generated by majority vote between all provided annotations labels of ``CV'' (Call for Violence) and ``HD'' (Human Degradation) which are two sub-types of hate speech. 
Final dataset include 2254 positive labels of hate among 27557 items. Secondly, \citet{de2018hate} provide an annotated corpus of English (\textbf{Storm}). We used posts included in ``all\_files'',\footnote{\url{https://github.com/Vicomtech/hate-speech-dataset}} and generated our own train and test subset. The final dataset includes 1196 positive labels among 10944 items.

For each dataset, the train and test set were split based on maintaining the same ratio of SGTs in both sets. Similarly, in each fold of cross validation 20\% of the train set was selected for validation purposes based on maintaining the same ratio of hate labels.

\noindent\textbf{Fairness Evaluation Datasets}

We used three out-of-domain datasets for evaluating fairness:
First, an existing dataset of stereotypes in English (``Dissimilar Counterfactuals''; DC) collected by \citet{nadeem2020stereoset} was applied, which contains two types of stereotype: \textit{intersentence} instances consisted of a base sentence provided for a target group and a stereotypical sentence generated by annotators for the same group, while \textit{intrasentence} instances were single sentences annotated as stereotypes.
For each sentence, we substitute the target group with all our SGTs, resulting in 25565 samples.

Second, ``Similar Counterfactuals'' (SC) consists of 77k synthetic English sentences generated by \citet{dixon2018measuring}. After removing sentences with less that 4 tokens, we ended up with 3200 sentences.

Third, following \citet{kennedy2020posthoc} we use a corpus of New York Times (NYT) articles to measure false positive rate. Specifically, for each SGT in our list (see Section~\ref{app:sgt}), we sampled 500 articles containing a mention of this SGT (and no other SGT mentions). This produced a balanced random sample of SGTs, which are heuristically assumed to have no hate speech (excepting rare occurrences, e.g., quotations).
    
\textbf{Evaluation}
For evaluating the Counterfactual Token Fairness (CTF) among a sentence and the list of its counterfactuals, we computed the cosine similarity of the 2D logits, produced as the output of \textbf{Bert-For-Sequence-Classification} model. We then calculated the average of these similarities to get a CTF value for the sentence and computed the average of CTFs over the dataset.

\textbf{Analysis of the Regularization Coefficient}
As mentioned in Section \ref{sec:clp}, the regularization coefficient $\lambda$ controls the extent to which counterfactual logit pairing formulation affects the training process. A larger value of $\lambda$ is expected to increases the importance of bias mitigation, while decreasing the essential classification performance. While in our experiments in Section \ref{sec:exp_hate} we set the same value for $\lambda$ for all counterfactual pairing approaches, here we discuss $\lambda$ as it creates a trade-off between classification accuracy and counterfactual token fairness.

Figure \ref{fig:ghc_lambda} and \ref{fig:storm_lambda} demonstrate the effect of $\lambda$ on the three main approaches evaluated on \textbf{GHC} and \textbf{Storm} datasets; 1) our approach for counterfactual generation (\textbf{CLP+ACL}), 2) \citet{garg2019counterfactual}'s approach which considers all counterfactuals of non-hate samples (\textbf{CLP+NEG}), and 3) counterfactual generation based on similar social categories (\textbf{CLP+SG}). As the plots denote, higher value of $\lambda$ corresponds with lower classification accuracy and lower (more desirable) counterfactual token fairness. These results also denotes that our proposed method \textbf{CLP+ACL}, achieves higher accuracy and fairness compared to the other approaches with different values of $\lambda$. 
In our experiments reported in Table~\ref{tab:results_ghc_storm}, $\lambda$ is set to 0.2 for all approaches that are based on counterfactual pairing (\textbf{CLP+}*). We chose this value, since based on the observed results in \ref{fig:ghc_lambda} and \ref{fig:storm_lambda}, it demonstrates the effect of counterfactual pairing loss on improving the fairness metrics while preserving classification accuracy. Future applications of our approach should rely on fine-tuning $\lambda$ during training.

\subfile{images/lambda_plot.tex}

\subsection{Glossary}
\textbf{Unintended bias}: When a model is biased with respect to a feature that it was not intended to be (e.g. race in Toxicity classifier).\\

\textbf{Group Fairness}: Fairness defintions that treat different groups equally (e.g. equality of odds, equality of opportunity.)\\

\textbf{Individual Fairness}: Fairness definitions that ensure similar predictions to similar individuals (e.g. counterfactual fairness.)\\

\textbf{Equality of Odds}:   ``A predictor $\hat{Y}$ satisfies equalized odds with respect to protected attribute $A$ and outcome $Y$, if $\hat{Y}$ and $A$ are independent conditional on $Y$. $P(\hat{Y}=1|A=0,Y =y) = P(\hat{Y}=1|A=1,Y =y) , y \in \{0,1\}$'', \cite{hardt2016equality}\\

\textbf{Equality of Opportunity}: ``A binary predictor $\hat{Y}$ satisfies equal opportunity with respect to $A$ and $Y$ if $P(\hat{Y}=1|A=0,Y=1) = P(\hat{Y}=1|A=1,Y=1)$'', \cite{hardt2016equality}\\

\textbf{Counterfactual}: Counterfactual conditionals are conditional sentences that assess the outcome under different circumstances. Here we use \cite{garg2019counterfactual} definition of counterfactual questions, ``How would the prediction change if the sensitive attribute referenced in the example were different?'' with SGT as the sensitive attribute\\

\textbf{Counterfactual reasoning}: The process of inferences from counterfactual conditionals compared to regular conditionals.\\

\textbf{Stereotype}: Stereotyping is a cognitive bias, deeply rooted in human nature \citep{cuddy2009stereotype} and omnipresent in everyday life through which humans can promptly assess whether an outgroup is a threat or not. 
Stereotyping, along with other cognitive biases, impacts how individuals create their subjective social reality as a basis for social judgements and behaviors \citep{greifeneder2017social}.
Stereotypes are often studied in terms of the associations that automatically influence judgement and behavior when relevant social categories are activated \citep{greenwald1995implicit}.\\

\subfile{tables/garg.tex}

%% file: images/lambda_plot.tex
\begin{figure*}[b!]
     \centering
     \begin{subfigure}[b]{0.3\textwidth}
         \centering
         \includegraphics[width=\textwidth]{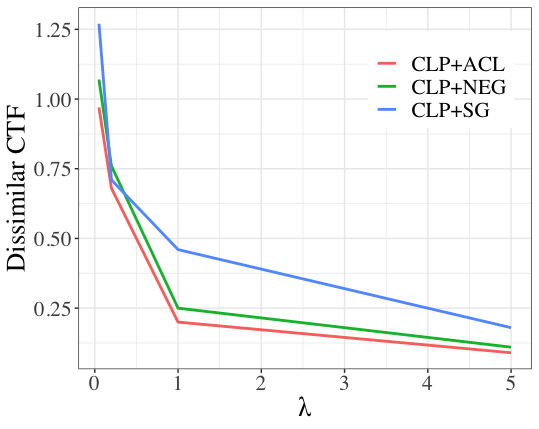}
     \end{subfigure}
     \hfill
     \begin{subfigure}[b]{0.3\textwidth}
         \centering
         \includegraphics[width=\textwidth]{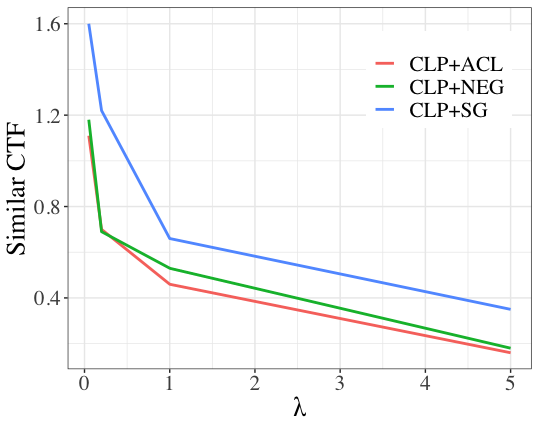}
     \end{subfigure}
     \hfill
     \begin{subfigure}[b]{0.3\textwidth}
         \centering
         \includegraphics[width=\textwidth]{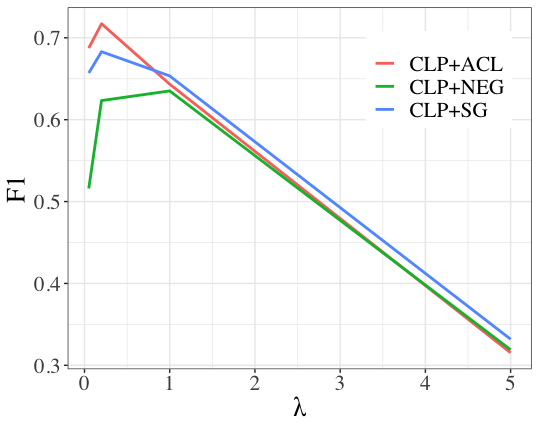}
     \end{subfigure}
        \caption{Changing the value of $\lambda$ while training models on the \textbf{GHC} dataset demonstrated the tradeoff between accuracy and counterfactual token fairness (evaluated on two datasets of dissimilar and similar counterfactuals).}
        \label{fig:ghc_lambda}
\end{figure*}

\begin{figure*}[b!]
     \centering
     \begin{subfigure}[b]{0.3\textwidth}
         \centering
         \includegraphics[width=\textwidth]{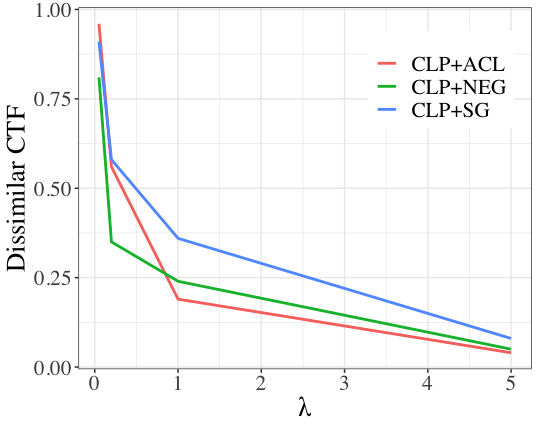}
     \end{subfigure}
     \hfill
     \begin{subfigure}[b]{0.3\textwidth}
         \centering
         \includegraphics[width=\textwidth]{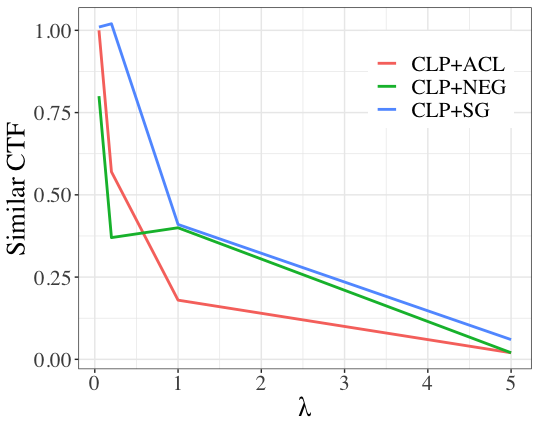}
     \end{subfigure}
     \hfill
     \begin{subfigure}[b]{0.3\textwidth}
         \centering
         \includegraphics[width=\textwidth]{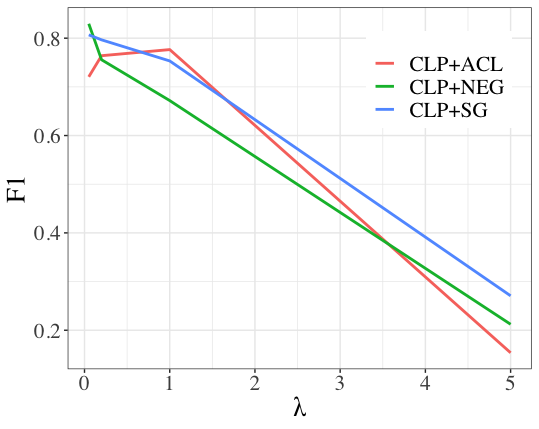}
     \end{subfigure}
        \caption{\change{Changing the value of $\lambda$ while training models on the \textbf{Storm} dataset demonstrated the tradeoff between accuracy and counterfactual token fairness (evaluated on two datasets of dissimilar and similar counterfactuals).}}
        \label{fig:storm_lambda}
\end{figure*}

%% file: tables/garg.tex
\begin{table*}[t]
    \centering
    \begin{tabular}{|c|p{.35\textwidth}|p{.35\textwidth}|}\toprule
        & \textbf{Non-hate Sample} & \textbf{Hate Sample} \\\hline\hline
        \cite{garg2019counterfactual} & All Counterfactuals & No Counterfactuals\\\hline
        Issues & Adding noisy synthetic data into the model since SGTs cannot interchangeably appear in all contexts &  Not supporting fairness for specific SGTs with high association with hate speech \citep{dixon2018measuring}\\\hline\hline
        Current approach &\multicolumn{2}{c|}{Counterfactuals with higher likelihood} \\\hline
        Improvement & Preventing counterfactuals with lower sentence likelihood, that can be noisy instances & Equalizing outputs for current instances and their more stereotypical counterfactuals\\\bottomrule
    \end{tabular}
    \caption{The through comparison of the proposed approach with \citet{garg2019counterfactual}, based on their solutions for positive and negative instances of hate speech}
    \label{tab:garg}
\end{table*}